\documentclass[11pt]{article}

\usepackage[utf8]{inputenc}
\usepackage[T1]{fontenc}
\usepackage[margin=1in]{geometry}
\usepackage{amsmath,amssymb,amsthm}
\usepackage{booktabs}
\usepackage{array}
\usepackage{graphicx}
\usepackage{float}
\usepackage{xcolor}
\usepackage[protrusion=true,expansion=false]{microtype}
\usepackage{enumitem}
\usepackage[numbers,square]{natbib}
\usepackage[colorlinks=true,linkcolor=blue,citecolor=blue,urlcolor=blue]{hyperref}
\usepackage{listings}

\title{\bfseries Teacher-Free Self-Training Amplifies but Does Not Compound:\\
	A Pass@$K$ Crossover on a Free-Verifier Domain}

\author{Igor Strozzi \\
	Applied Mathematics Department \\
	Federal University of Rio de Janeiro}
\date{Preprint. \today}

\begin{document}
	\maketitle
	
	\begin{abstract}
		When a language model trains on its own verified outputs, does it \emph{acquire}
		capability beyond its base, or merely get better at \emph{expressing} capability
		the base already had? We make the question decidable with a teacher-free
		``constellation''---a generator, a learned critic, and a free exact
		verifier---on a FlashFill-style ``trapdoor'' DSL, where verified (problem,
		solution) pairs are cheap to synthesize, hard to invert, and free to check
		exactly. Everything runs on one 4-bit Qwen3-4B on a single 24~GB GPU, with no
		model in the loop larger than the base. We report three findings.
		(i)~Critic-guided selection beats verifier-filtered best-of-$k$ by $+9.1$~pp
		($6/6$ seeds), with the entire gain localized to tasks where candidates disagree
		on held-out inputs. (ii)~Per-round STaR self-training raises the ceiling but
		never accelerates---the gain tracks remaining headroom and decelerates across
		$K{=}4$ independent training trajectories. (iii)~The domain has no clean
		zero-capability frontier, so the usual ``$0\%\!\to\!$climb $=$ emergence'' test
		is invalid here. A measured pass@$K$ crossover settles the diagnosis: the
		trained model wins at the operating budget (pass@8) but the base \emph{overtakes}
		it at a large budget (pass@$64$) on every trajectory, so self-training
		\emph{concentrates} probability mass rather than expanding reach. This is
		amplification, not compounding. ($K{=}4$ is indicative, not yet a robust
		across-trajectory CI.)
	\end{abstract}
	
	\section{Introduction}
	Self-improvement loops for language models raise a question that is easy to
	state and surprisingly hard to answer cleanly. When a system trains on its own
	verified outputs, does it \emph{compound} capability---following a trajectory
	whose per-round returns are self-reinforcing and that eventually lifts the model
	past the reach of its own base distribution---or does it merely \emph{amplify},
	sharpening and surfacing capability the base could already reach given enough
	sampling? Two recurring habits blur the distinction. First, most loops place a
	stronger teacher somewhere in the chain (a larger model, a reward model, a
	human), which caps any apparent novelty at the teacher's level by construction,
	so one can never tell whether the loop itself produced the gain. Second,
	``emergence'' is routinely declared whenever a metric moves off an apparent
	zero, even when that zero is a sampling artifact rather than a genuine inability.
	
\begin{figure}[h]\centering
		\includegraphics[width=0.80\linewidth]{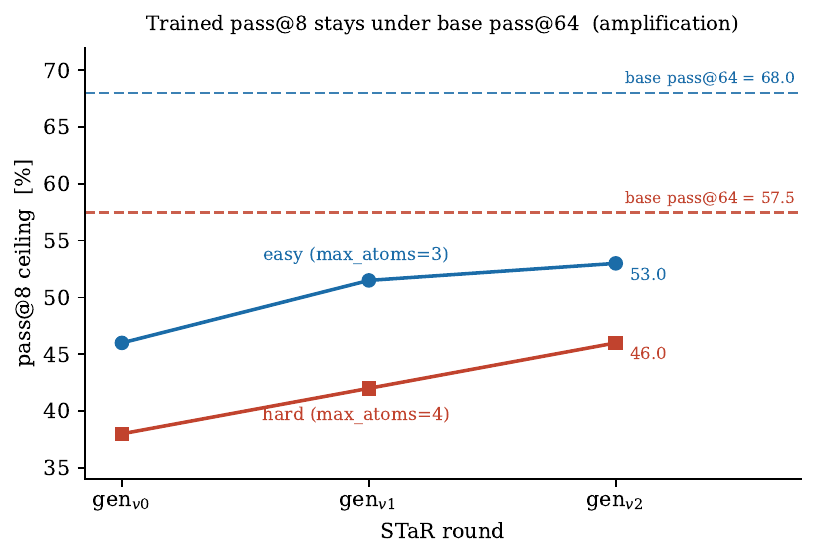}
		\caption{The measurement that adjudicates the headline, now populated. Solid
			markers are \emph{measured} pass@8 ceilings of $v_0/v_1/v_2$ (seed 1234) on the
			easy (circles) and hard (squares) banks; dashed lines are the base $v_0$
			pass@$64$ ceiling per bank---a lower bound on the reachable frontier. On both
			banks the rising pass@8 trajectory stays \emph{under} the base pass@$64$ line
			(easy $53.0\%$ vs $68.0\%$; hard $46.0\%$ vs $57.5\%$): self-training surfaces
			reachable-but-rare programs without breaking above base reach---amplification,
			not compounding. Because pass@$64$ lower-bounds $K\!\to\!\infty$, the gap is
			conservative.}
		\label{fig:envelope}
	\end{figure}

	Our first move is to make the distinction operational rather than rhetorical.
	Following what has become standard practice in the elicitation-versus-acquisition
	literature \citep{yue2025rlvr,he2025unlikely}, we take the base model's pass@$K$
	curve at \emph{large} $K$ to be its reachable frontier---the set of problems it
	can solve if allowed enough independent attempts---and define the two outcomes
	against it:
	\begin{itemize}[leftmargin=1.4em,itemsep=2pt]
		\item \textbf{amplification}: per-round improvement that is monotone but
		non-accelerating, and whose ceiling at the operating budget stays
		\emph{under} the base model's large-$K$ envelope. The loop is surfacing
		reachable-but-rare programs, not new ones.
		\item \textbf{compounding}: per-round gain that accelerates, or a ceiling that
		\emph{breaks above} the base large-$K$ envelope---genuinely new
		reachability, of the kind prolonged RL is reported to produce
		\citep{liu2025prorl}.
	\end{itemize}
	We want to be clear about what we are and are not claiming. The \emph{level} of
	the ceiling does rise across rounds in our experiments; self-training visibly
	helps. Our contribution is that the rise is headroom-paced and non-accelerating,
	and---by the measured pass@$K$ crossover of Fig.~\ref{fig:envelope}, in which the
	base overtakes the trained model at large $K$ on every training trajectory---it
	is amplification, not compounding.
	
	The second move is the choice of domain. We isolate the question with a
	\emph{``trapdoor domain''}: a problem class in which forward generation of
	\emph{verified} (problem, solution) pairs is cheap, backward solving is hard for
	the model, and verification is a free, total, and exact function. A single
	object---an exact interpreter for a small string-transformation language---then
	plays three roles at once. It is the verifier; it is the engine of a
	teacher-free data-synthesis pipeline; and, because no larger model's judgment
	ever enters the loop, it is what keeps genuine novelty \emph{possible} rather
	than capping it at a teacher's competence. That last point is what lets a
	negative result here mean something: if compounding were going to appear, the
	construction would not have prevented it.
	
	\paragraph{Contributions.}
	\begin{enumerate}[leftmargin=1.6em,itemsep=2pt]
		\item A fully teacher-free generator/critic/verifier constellation on a
		trapdoor string-transformation DSL, runnable on a single 24~GB GPU,
		with every component and analysis instrument unit-tested.
		\item Evidence that critic-guided selection beats verifier-filtered
		best-of-$k$, together with a consistency check that localizes the entire
		gain to the structurally-relevant subset---the tasks where surviving
		candidates disagree on held-out inputs.
		\item A three-round, two-distribution self-training trajectory that exhibits
		headroom-paced amplification without acceleration or collapse, supported
		by a confound-resistant gradient metric. The metric is
		headroom-normalized (so a low-base axis is not credited for climbing
		faster simply because it started lower), low-base-controlled (it carries
		a built-in control axis at comparable starting height), and same-seed
		paired (so per-round gain is separated from evaluation-split noise).
		\item A measured pass@$K$ crossover that confirms the amplification diagnosis:
		the trained model wins at the operating budget but the base overtakes it
		at large $K$, robustly across four independent training trajectories on
		both task distributions (Fig.~\ref{fig:envelope}).
		\item A negative methodological finding: in a compositional DSL, task cells
		that read zero at small sample counts are undersampling artifacts rather
		than capability frontiers, so the naive ``$0\%\!\to\!$climb $=$
		emergence'' test is invalid here. This is a direct reply to
		existence-of-compounding claims grounded in zero-pass@$k$ cells
		\citep{liu2025prorl}.
	\end{enumerate}
	
	\section{Related work}\label{sec:related}
	\paragraph{Elicitation vs.\ acquisition---the question we inherit.}
	Whether training on verified outputs \emph{acquires} capability beyond the base
	model or merely \emph{elicits} latent capability is the central open question of
	the current RLVR literature, and it is precisely our amplification-versus-compounding
	distinction in a different regime. \citet{yue2025rlvr} make the strongest
	elicitation case: across model families, RL algorithms, and math/code/visual
	benchmarks, RL-trained models beat their base at low $k$ (pass@1) but are
	\emph{overtaken} by the base at large $k$, with the boundary of solvable
	problems \emph{narrowing} as training proceeds; their coverage and perplexity
	analysis places the successful RL traces inside the base model's own sampling
	distribution. The methodological lesson we take from them is the instrument
	itself---that the hypotheses are separated by pass@$k$ at large $k$, not by
	pass@1 (\S\ref{sec:gradient}, Fig.~\ref{fig:envelope}).
	
	\paragraph{The counter-case.}
	\citet{liu2025prorl} (ProRL) contest the elicitation thesis on its own metric.
	With KL control, reference-policy resetting, and a diverse task suite, prolonged
	RL raises pass@$k$ across the whole range and solves instances the base fails at
	\emph{any} budget ($0\%$ pass@$k$), including out-of-training-size generalization
	on a graph-coloring task. This is the clearest existence proof of compounding,
	and it is the direct foil for two of our results: the no-zero-frontier finding
	(\S\ref{sec:nozero}) and our out-of-training atom-count generalization on the
	hard bank. We do \emph{not} observe the ProRL phenomenon, and we argue the
	difference is regime-bound---teacher-free SFT-STaR with a fixed base and no
	exploration bonus, a few rounds rather than thousands of steps---and possibly
	induced by our free exact verifier, which only ever reinforces
	correct-and-reachable programs and so has no way to push outward.
	
	\paragraph{Self-training / bootstrapping.}
	Our loop sits in the STaR \citep{zelikman2022star} / rejection-sampling-FT
	\citep{yuan2023rft} / ReST \citep{gulcehre2023rest} family: keep
	model-generated, verifier-confirmed outputs as supervised fine-tuning targets,
	then retrain. What sets our instance apart is that it is teacher-free by
	construction---the verifier is a total exact interpreter, not a model or a
	reward net---so the ceiling cannot be a teacher's competence in disguise. This
	matters in light of \citet{yue2025rlvr}'s finding that distillation, unlike RL,
	\emph{does} import a teacher's reasoning patterns; we have removed that channel
	entirely.
	
	\paragraph{Diversity collapse and our no-collapse result.}
	A growing line of work formalizes why accuracy-optimizing post-training collapses
	generation diversity. \citet{dang2025weight} document pass@$k$ deterioration
	under SFT and self-improvement, give a bias--variance decomposition of expected
	pass@$k$, and mitigate it via weight interpolation (WiSE-FT); differential
	smoothing \citep{diffsmooth2025} supplies a selection/reinforcement-bias account
	together with a reward fix; \citet{he2025unlikely} trace RL's weak multi-sample
	performance to an inability to reinforce low-probability correct samples; and
	\citet{jointdiv2025} optimize diversity and quality jointly. Against this
	backdrop, our observation that STaR-data diversity \emph{rises} while the pass@8
	ceiling does not regress is a contrasting data point---with one caveat we state
	plainly: our diversity metric is solution-set cardinality under a free verifier,
	not the entropy or $n$-gram diversity these works track, so the comparison is
	qualitative rather than a refutation.
	
	\paragraph{Program synthesis from examples.}
	The domain is inductive programming-by-example in the FlashFill lineage
	\citep{gulwani2011flashfill}. Our contribution there is not a synthesizer but
	the use of a free exact verifier as a teacher-free reward.
	
	\paragraph{Positioning.}
	We land on the elicitation side of \citet{yue2025rlvr}, but in a regime that
	line has not examined: teacher-free iterated SFT on a domain with a free, total,
	exact verifier, where ``reachable'' can in principle be measured exactly via
	large-$K$ pass@$K$ and the no-teacher construction removes the distillation
	escape hatch. The combination we report---no compounding \emph{and} no
	collapse---is one that neither pole of the debate predicts for a single run.
	
	\paragraph{What we baseline against.}
	Each of our three claims is internally baselined against the correct published
	referent, so that no external benchmark is silently missing. Critic-guided
	selection is compared to verifier-filtered best-of-$k$ (``naive-pick''), which
	is best-of-$n$ selection against a verifier \citep{cobbe2021verifiers}---the
	standard baseline a learned selector must beat, and which our critic beats on
	$6/6$ seeds. The self-training trajectory \emph{is} faithful STaR
	\citep{zelikman2022star}: verifier-only rejection sampling on the model's own
	generations, retrain, repeat, with the critic confined to eval-time selection
	and kept out of the training loop. The method is therefore the published method,
	and the trajectory is itself the baseline. Finally, the
	amplification-versus-compounding adjudication is baselined against the base
	model's own pass@$K$ curve at large $K$ \citep{yue2025rlvr}, measured on a public
	artifact (Qwen3-4B). No external PBE accuracy number is missing: each claim is
	relative by construction and is compared to the standard referent for its own
	question.
	
	\section{The trapdoor domain}\label{sec:domain}
	\subsection{DSL and verifier}
	The substrate is a small language of FlashFill-style string transforms, with the
	following grammar:
	\begin{lstlisting}
		Expr := Concat([Atom, ...])
		Atom := ConstStr(s) | SubStr(Pos, Pos) | Case(op, Atom)   op in {upper,lower,cap}
		Pos  := CPos(k)                  # absolute index, k may be negative
		| MatchPos(tok, n, side)   # nth match of tok, start|end boundary
		tok  in {WORD, DIGITS, UPPER, LOWER}
	\end{lstlisting}
	The interpreter \texttt{run(prog, s)} is \emph{total or raises}: an out-of-range
	position raises \texttt{PosError} rather than silently clamping into bounds. This
	is a deliberate design choice. Clamping would let two programs that agree on the
	visible examples but diverge elsewhere look identical, and that divergence is
	exactly the signal we want to preserve. Correctness is decided by exact string
	equality, so the verifier has zero training cost and zero learned opinion. As a
	sanity guarantee on the representation, a canonical text renderer and its exact
	inverse parser are round-trip-proven: $0$ structural and $0$ behavioral failures
	over $5000$ sampled programs.
	
	\subsection{Trapdoor task generation}
	A task is built forward, from a known program. We sample a random program, run
	it on random structured inputs (names, dates, emails, codes), reject degenerate
	outputs (empty or identity), and keep tasks that yield at least $N$ distinct
	examples. Two measured properties make the resulting tasks usable. The generator
	has ${\sim}70\%$ non-degenerate yield, so producing data is cheap; and tasks are
	\emph{discriminative} at the $0.5\%$ level---a random \emph{different} program
	satisfies a task's visible examples only $0.5\%$ of the time ($5$ spurious solves
	over $n{=}1000$ probes, namely $200$ tasks $\times\,5$ random alternative programs
	each). A task therefore pins down its program tightly enough to carry real signal
	rather than admitting many trivial fits.
	
	\subsection{Structural task classes}
	For analysis only---never as a training signal---each program is bucketed by a
	deterministic structural signature: its atom count plus its set of flags
	(\texttt{sub}, \texttt{const}, \texttt{match}, \texttt{case}, \texttt{neg}),
	e.g.\ \texttt{a3+sub+const+match+neg}. The gradient study of \S\ref{sec:gradient}
	uses four overlapping analysis axes drawn from these signatures:
	\texttt{const-anchored}, \texttt{match-without-const} (which we call the weak
	axis), \texttt{neg-bearing}, and \texttt{case-bearing}.
	
	\section{Method}\label{sec:method}
	\subsection{Roles}
	The system is a constellation of four roles, three of which share a single base
	model through swappable adapters:
	\begin{itemize}[leftmargin=1.4em,itemsep=2pt]
		\item \textbf{Verifier} --- \texttt{dsl.run}. Total, exact, and free; it
		carries no parameters and no learned judgment.
		\item \textbf{Generator} --- a 4B LoRA adapter that maps
		\texttt{visible examples} $\to$ \texttt{program}.
		\item \textbf{Critic} --- a 4B LoRA adapter implementing a generative yes/no
		classifier, $(\text{examples}, \text{candidate}) \to
		\text{generalizes?}$, trained on the generator's
		\emph{verifier-labeled} candidates and used at inference by ranking
		candidates on the $P(\text{yes}) - P(\text{no})$ logit margin. The
		reason a learned critic is needed at all is subtle: at selection time
		only the \emph{visible} examples are available, so the verifier can
		confirm that a candidate is consistent with what is visible, but it
		cannot adjudicate held-out generalization until \emph{after} a pick has
		been committed. The critic is what fills that gap.
		\item \textbf{Conductor} --- model-free; it holds the blackboard state and the
		termination rule.
	\end{itemize}
	
	\subsection{The loop}
	A single pass through the system proceeds in five steps. (1)~Split each task's
	examples into a \textbf{visible} set and a \textbf{held-out} set. (2)~The
	generator proposes $k$ candidate programs from the visible examples. (3)~The
	verifier discards any candidate that fails a visible example---free and total,
	so this costs nothing. (4)~The critic ranks the survivors and the argmax is
	selected. (5)~The verifier adjudicates the chosen candidate on the held-out
	examples, which is what tells us whether the pick truly generalized.
	
	\subsection{Teacher-free self-training (STaR)}
	Self-training proceeds in rounds. In each round we run the current generator over
	fresh, non-bank tasks and keep \emph{only the model-generated, verifier-confirmed
		generalizing programs} as new supervised fine-tuning targets, then train the next
	generator on them. The secret program behind a task is used only to label the
	task by structural axis; it is \emph{never emitted as a target} (a property
	checked by \texttt{smoke\_roundlogic}). This is precisely what makes any gain
	self-bootstrapped rather than quietly distilled from an oracle. We deliberately
	run \emph{pure per-round} STaR---each generator is trained on that round's data
	alone, with no cumulative anchor to earlier rounds---both to make any collapse
	maximally visible and to hold the method fixed across the trajectory.
	
	\subsection{Confound-resistant gradient metric}\label{sec:gradient}
	To ask whether self-training improved the weak axis \emph{preferentially}, we
	cannot simply compare raw percentage-point gains: an axis that starts lower will
	climb more in raw points at equal effort, so a raw comparison is confounded by
	starting height. We instead report the \emph{fraction of available ceiling
		headroom closed} per axis,
	\[
	\text{headroom-closed} \;=\; \frac{\Delta_{\text{rate}}}{1 - \text{rate}_0},
	\]
	which normalizes each axis by how much room it had to improve. To guard against
	the residual worry that any low-base axis improves more, the \texttt{neg-bearing}
	axis serves as a built-in \emph{low-base control}: only if the weak axis closes
	more headroom than \emph{both} the easy axis \emph{and} the comparably-low-base
	\texttt{neg} control can the effect be attributed to weakness rather than to
	generic low-base regression. Separately, the base model's large-$K$ pass@$K$
	envelope---measured on its own---is what ultimately adjudicates amplification
	versus compounding: the trained model's ceiling stays under it
	(Fig.~\ref{fig:envelope}, \S\ref{sec:limits}).
	
	\section{Experimental setup}\label{sec:setup}
	\begin{itemize}[leftmargin=1.4em,itemsep=2pt]
		\item \textbf{Base model:} Qwen3-4B-Instruct-2507 \citep{qwen3}, 4-bit
		\texttt{nf4} quantization, \texttt{bf16} compute.
		\item \textbf{Adapters:} LoRA $r{=}16$, $\alpha{=}32$, dropout $0.05$,
		\texttt{all-linear} targets (${\sim}0.81\%$ trainable parameters);
		QLoRA with \texttt{paged\_adamw\_8bit}.
		\item \textbf{Hardware:} a single RTX 3090 (24~GB). The generator and critic
		adapters run as two sequential passes over one quantized base, with the
		generator freed between passes---the two adapters cannot co-reside in
		full precision at this budget.
		\item \textbf{Frozen eval banks} (never trained on, sha-pinned, with
		tamper-evident metadata): \emph{easy} (\texttt{max\_atoms=3}, $200$
		tasks, sha \texttt{512ddf40\dots}); \emph{hard} (\texttt{max\_atoms=4},
		$200$ tasks, sha \texttt{49807fcc\dots}, an out-of-training atom count).
		Each task carries a $12$-example pool, with the visible/held-out split
		carved at eval time.
		\item \textbf{Default regime:} visible$=2$, $8$ generator samples per task---the
		divergence-maximizing peak of the Stage-0 sweep (\S\ref{sec:stage0}).
		\item \textbf{Software} (probe-confirmed at run time): torch 2.12+cu130,
		transformers 5.9, trl 1.5.1, peft 0.19.1, bitsandbytes 0.49.2,
		Python 3.13.
	\end{itemize}
	
	\section{Results}\label{sec:results}
	Throughout, raw counts are primary and rates are derived; all denominators are
	$/200$ unless noted otherwise.
	
	\subsection{Stage 0 --- generator baseline and regime sweep}\label{sec:stage0}
	Before adding a critic or any self-training, we characterize the bare generator
	and choose an operating point. The free variable is how many visible examples the
	generator is shown; Table~\ref{tab:sweep} sweeps it.
	\begin{table}[h]\centering
		\caption{Stage-0 regime sweep (easy bank, seed 1234).}\label{tab:sweep}
		\begin{tabular}{ccccc}
			\toprule
			visible & solve@1 & naive best-of-$k$ & verifier ceiling & divergence \\
			\midrule
			1 & 14.5\% & 17.5\% & 31.0\% & 22.5\% \\
			\textbf{2} & \textbf{25.5\%} & \textbf{31.5\%} & \textbf{46.0\%} & \textbf{26.5\%} \\
			3 & 28.5\% & 35.0\% & 46.0\% & 17.5\% \\
			5 & 34.5\% & 49.5\% & 52.5\% & 10.5\% \\
			\bottomrule
		\end{tabular}
	\end{table}
	
	The quantity to watch is \emph{divergence}---the fraction of tasks on which the
	surviving candidates disagree, which is the only place a selector can do any
	work---and it is humped in the number of visible examples. At visible$=2$,
	divergence is maximized while the ceiling is already tied with visible$=3$; by
	visible$=5$, naive best-of-$k$ has nearly caught the ceiling and a selector has
	almost nothing left to do. We therefore fix visible$=2$ as the operating point
	for all subsequent stages: it is where a learned selector has the most room to
	help.
	
	\subsection{Stage 0 --- six-seed confidence interval}\label{sec:stage0ci}
	Single-seed rates carry split noise, so we re-measure the baseline across six
	evaluation-split seeds (Table~\ref{tab:ci}) to see which quantities are stable
	properties of the model and which are draw-dependent.
	\begin{table}[h]\centering
		\caption{Six-seed CI for gen\_v0 (easy bank, visible$=2$), seeds
			$\{1234,7,99,256,4242,8888\}$. Mean / sample-stdev (ddof$=1$) / min / max
			across seeds.}\label{tab:ci}
		\begin{tabular}{lccccc}
			\toprule
			metric & per-seed $(/200)$ & mean & stdev & min & max \\
			\midrule
			solve@1 (greedy)       & 51,48,50,40,53,51   & 24.4\% & 2.3\% & 20.0\% & 26.5\% \\
			naive best-of-$k$      & 71,60,77,62,72,67   & 34.1\% & 3.2\% & 30.0\% & 38.5\% \\
			\textbf{verifier ceiling} & 92,91,99,90,96,92 & \textbf{46.7\%} & \textbf{1.7\%} & 45.0\% & 49.5\% \\
			any visible-consistent & 118,114,125,113,122,121 & 59.4\% & 2.4\% & 56.5\% & 62.5\% \\
			divergent              & 47,54,40,54,45,49   & 24.1\% & 2.7\% & 20.0\% & 27.0\% \\
			\bottomrule
		\end{tabular}
	\end{table}
	
	The verifier ceiling is the tightest metric ($\sigma{=}1.7\%$), which tells us
	the generator's reachable frontier is a stable model property rather than an
	artifact of any particular split; single-seed absolute rates elsewhere carry
	roughly $\pm 2$--$3$~pp of one-draw split noise. (This CI covers the baseline
	only; see \S\ref{sec:limits}.) One apparent discrepancy is worth resolving
	directly. The seed-1234 naive best-of-$k$ differs between the Stage-0 sweep
	(\S\ref{sec:stage0}, $63/200$) and this CI's seed-1234 entry ($71/200$) at the
	identical visible$=2$, $k{=}8$ setting. The cause is benign: candidate generation
	is stochastic (\texttt{do\_sample} with no fixed generation seed), so the two
	runs simply draw different candidate sets. The deterministic-greedy solve@1
	reproduces exactly ($51/200$ in both), which confirms the difference is
	best-of-$k$ sampling variance rather than a stale figure, and both values lie
	inside the six-seed naive best-of-$k$ spread ($60$--$77$) reported here. For
	absolute rates, the single-draw sweep is superseded by this seed-resolved CI.
	
	\subsection{Stage 1 --- critic-guided selection beats naive}\label{sec:stage1}
	We now add the critic and ask whether learned selection beats the naive
	verifier-filtered baseline (Table~\ref{tab:critic}).
	\begin{table}[h]\centering
		\caption{Critic vs.\ naive selection (visible$=2$), two split seeds.}\label{tab:critic}
		\begin{tabular}{lcc}
			\toprule
			& seed 1234 & seed 7 \\
			\midrule
			Overall naive-pick      & 67 (33.5\%) & 68 (34.0\%) \\
			Overall \textbf{critic-pick} & \textbf{85 (42.5\%)} & \textbf{81 (40.5\%)} \\
			Verifier ceiling        & 88 (44.0\%) & 84 (42.0\%) \\
			Divergent naive-pick    & 24/45 (53.3\%) & 26/42 (61.9\%) \\
			Divergent \textbf{critic-pick} & \textbf{42/45 (93.3\%)} & \textbf{39/42 (92.9\%)} \\
			\bottomrule
		\end{tabular}
	\end{table}
	
	Critic-pick beats naive-pick on both seeds, reaching about $93\%$ on the
	divergent subset. The most informative check is one of internal consistency: the
	overall improvement equals the divergent-subset improvement \emph{exactly}
	(seed 1234: $+18 = +18$; seed 7: $+13 = +13$). The critic, in other words, moved
	outcomes only on the tasks where the survivors disagree on held-out inputs---which
	is exactly, and only, where a selector can structurally make a difference. This
	is the signature of amplification: a selector recovering capability already
	present and capped by the generator's ceiling, not the creation of new
	capability.
	
	\paragraph{Six-seed CI.} Re-running the loop across the six seeds, the headline
	quantity is the within-seed paired delta between critic and naive selection (with
	a fixed denominator $T{=}200$): the per-seed deltas are
	$[18,13,18,25,17,18]$, giving a mean of $+9.1\%$ ($\sigma\,1.9\%$), a range of
	$+6.5\%$--$+12.5\%$, and a positive sign on all $6/6$ seeds. On the divergent
	subset---where the per-seed denominator varies, so the per-seed rate is the unit
	of replication---the critic's rate-of-rates is $92.5\%$ ($\sigma\,1.9\%$) against
	naive selection's $53.3\%$ ($\sigma\,7.6\%$). The critic's divergent-subset
	advantage is the single most seed-stable quantity in the project, and it removes
	exactly the split-dependence that makes naive selection noisy.
	
	\subsection{Stage 2 --- self-training trajectory}\label{sec:stage2}
	With selection characterized, we turn to the central question: what happens to
	the generator's own ceiling as it trains on its verified output across rounds?
	Table~\ref{tab:traj} tracks the verifier ceiling for $v_0 \to v_1 \to v_2$ on
	both banks.
	\begin{table}[h]\centering
		\caption{Verifier-ceiling trajectory as six-seed marginal CIs (mean rate,
			$\sigma$ across seeds $\{1234,7,99,256,4242,8888\}$).}\label{tab:traj}
		\begin{tabular}{lccc}
			\toprule
			bank & v0 & v1 & v2 \\
			\midrule
			easy (\texttt{max\_atoms=3}) & 46.7\% ($\sigma$1.7) & 53.1\% ($\sigma$3.0) & 55.4\% ($\sigma$1.7) \\
			hard (\texttt{max\_atoms=4}) & 37.9\% ($\sigma$2.6) & 41.1\% ($\sigma$1.1) & 45.0\% ($\sigma$1.4) \\
			\bottomrule
		\end{tabular}
	\end{table}
	
	\begin{figure}[h]\centering
		\includegraphics[width=0.74\linewidth]{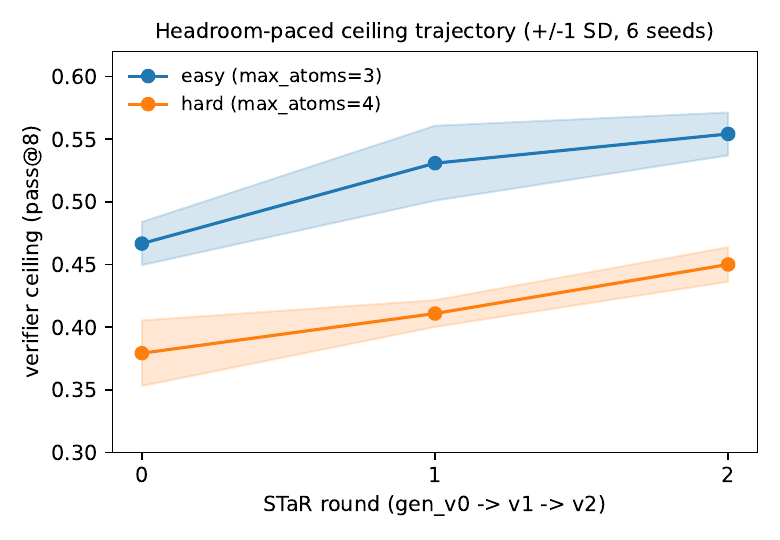}
		\caption{Verifier-ceiling (pass@8) vs.\ STaR round. Bands are ${\pm}1\sigma$
			across six eval-split seeds; annotated $\Delta$ are within-seed paired
			per-round deltas (mean, $\sigma$, sign tally). Per-round gain decelerates on
			the high-base easy bank and holds on the low-base hard bank.}
		\label{fig:traj}
	\end{figure}
	
	The ceiling rises at every round on both banks, so self-training is clearly
	helping. The question is the \emph{shape} of that rise. Because the
	adjacent-round marginal CIs overlap, we test the per-round gain in paired form
	(the same split seeds at each round), shown in Table~\ref{tab:delta}.
	\begin{table}[h]\centering
		\caption{Paired per-seed ceiling delta (mean$\Delta$ over $T{=}200$, $\sigma$,
			sign tally $+/-$).}\label{tab:delta}
		\begin{tabular}{lcc}
			\toprule
			step & easy bank & hard bank \\
			\midrule
			v0$\to$v1 & $+6.4\%$ ($\sigma$3.0, 6/0) & $+3.2\%$ ($\sigma$2.8, \textbf{5/1}) \\
			v1$\to$v2 & $+2.3\%$ ($\sigma$2.4, 5/0) & $+3.9\%$ ($\sigma$0.9, 6/0) \\
			\bottomrule
		\end{tabular}
	\end{table}
	
	On this single ($t_0$) trajectory the easy bank decelerates cleanly
	($+6.4 \to +2.3$), but the hard-bank pair is, on its face, a mild
	\emph{increase} ($+3.2 \to +3.9$)---the one cell that could be read as
	acceleration, and therefore as compounding. Rather than explain it away, we test
	it directly with train-seed replication, which resolves it below.
	
	\paragraph{Train-seed robustness (Tier 3).} The eval-split CIs above vary only
	the evaluation split of \emph{fixed} adapters; they do not test whether the
	\emph{shape} of the trajectory survives a different self-training run. To check
	that, we re-ran the entire STaR pipeline as $K{=}4$ independent training
	trajectories (distinct training seeds, all evaluated at the fixed split seed
	1234) and report the across-trajectory per-round ceiling delta in
	Table~\ref{tab:tier3}.
	\begin{table}[h]\centering
		\caption{Across-trajectory ceiling delta over $K{=}4$ independent training
			runs (mean$\Delta$ over $T{=}200$, $\sigma$ across trajectories, sign tally).}\label{tab:tier3}
		\begin{tabular}{lcc}
			\toprule
			step & easy bank & hard bank \\
			\midrule
			v0$\to$v1 & $+7.2\%$ ($\sigma$1.8, 4/0) & $+5.1\%$ ($\sigma$1.7, 4/0) \\
			v1$\to$v2 & $+1.9\%$ ($\sigma$2.8, 3/1) & $+2.8\%$ ($\sigma$1.3, 4/0) \\
			\bottomrule
		\end{tabular}
	\end{table}
	
	Across independent training trajectories the no-acceleration pattern holds on
	\emph{both} banks: every trajectory decelerates from v0$\to$v1 to v1$\to$v2
	(easy $+7.2 \to +1.9$; hard $+5.1 \to +2.8$). The single-seed hard-bank
	``increase'' ($+3.2 \to +3.9$) was thus a $t_0$-trajectory artifact---it does not
	survive train-seed replication, and no trajectory shows acceleration. The
	asymmetry in the second step is itself on-message for ceiling exhaustion: on the
	near-ceiling easy bank the round-2 gain has \emph{essentially vanished}
	($+1.9\%$, $\sigma\,2.8$, $3/1$---one trajectory negative, indistinguishable from
	zero), whereas the low-base hard bank, with more reachable headroom still
	available, continues to gain unanimously ($+2.8\%$, $4/0$). We are careful not to
	overclaim: $K{=}4$ is not a robust confidence interval (\S\ref{sec:limits}). But
	the no-acceleration claim is no longer resting on a single observed trajectory.
	
	\paragraph{No collapse.} Self-training did not degrade the sampler. The
	STaR-data \texttt{diversity\_mean} (distinct generalizing programs per solved
	task) \emph{rose} from $2.79$ to $4.05$; the \texttt{any-visible-consistent}
	rate rose on both banks (paired $+3.8\%/+2.3\%$ on easy, $+2.2\%/+4.6\%$ on
	hard); and the statistically-solid easy class \texttt{a1+const} ($n{=}34$) never
	regressed. Where the gains shrink---on the easy bank---it is exhaustion of
	\emph{reachable} headroom under a fixed base, not degeneration of the sampler.
	
	\subsection{No clean zero-capability frontier exists in this DSL}\label{sec:nozero}
	A natural worry about any emergence claim is whether the apparent zeros are real.
	Here they are not. Trained only on \texttt{max\_atoms}${\leq}3$, gen\_v0 already
	solves unseen 4-atom classes on the hard bank at $25$--$50\%$ ceiling
	(for example \texttt{a4+sub+const+match+case} at $50\%$ and
	\texttt{a4+sub+const+match+neg} at $44\%$)---comparable to the same token-sets at
	atom-$3$. The \texttt{Concat} structure is compositional enough that ``one more
	atom'' is simply not out-of-distribution. The only cells that read zero are
	small-$n$ ($0/3$--$0/5$), and they are \emph{undersampling artifacts}: the same
	classes solve at $5$--$38\%$ once measured at population scale. Declaring
	emergence because such a cell climbs off zero would be mistaking variance for
	capability---which is exactly why we rely on the relative-gradient test of
	\S\ref{sec:gradient} rather than a naive zero-floor test. This speaks directly to
	existence-of-compounding claims that rest on zero-pass@$k$ cells
	\citep{liu2025prorl}: in a compositional domain, such cells must first be shown
	non-trivial at population scale.
	
	\paragraph{The real wall is above the ceiling, not at zero.} The genuine bound on
	this paradigm is not a zero-frontier but a gap higher up. At $K{=}64$ the base
	produces a visible-consistent (possibly overfit) program for $79.0\%$ (easy) /
	$68.5\%$ (hard) of tasks, yet only $68.0\%$ / $57.5\%$ ever yield a
	\emph{generalizing} one---a stable ${\sim}11$~pp ($22$-task) gap on both banks, a
	band in which the base reliably finds a fitting-but-wrong program and never the
	correct one. That gap, not a zero-frontier, is the real generator-capacity bound:
	the region no amount of selection or self-training over a fixed base can enter,
	and hence the ceiling on what this paradigm can deliver.
	
	\section{Discussion}\label{sec:discussion}
	\paragraph{Amplification, not (observed) compounding.} Every positive movement we
	saw traces back to surfacing reachable-but-rare capability---the selector
	recovering headroom, and STaR enriching the sampling tail. Nothing ever climbed
	off a population-scale zero. It is worth emphasizing that the teacher-free
	construction kept genuine novelty \emph{possible}; the finding, over three
	rounds, is simply that it did not \emph{appear}. The crossover is the full
	\citet{yue2025rlvr} signature, now measured on both sides (pass@$K$ at seed
	1234): $v_2$ beats the base at the operating budget (pass@8 $53.0\%/46.0\%$ vs
	base $46.0\%/38.0\%$, easy/hard) but the base \emph{overtakes} $v_2$ at a large
	budget (pass@64 base $68.0\%/57.5\%$ vs $v_2$ $61.5\%/52.5\%$). The overtaking is
	train-seed robust---it holds for all four independent training trajectories on
	both banks ($4/4$; \S\ref{sec:limits}). We therefore \emph{confirm} the large-$K$
	coverage narrowing rather than escaping it: self-training concentrates
	probability mass, making rare-but-reachable solutions common (so it wins at
	$k{=}8$) while \emph{lowering} large-$K$ coverage (so it loses at $k{=}64$). The
	within-model $K$-scaling makes the concentration concrete: raising the sample
	budget from $8$ to $64$ adds $+22$~pp of ceiling for the base ($46 \to 68$ on
	easy) but only $+8.5$~pp for $v_2$ ($53 \to 61.5$). The trained model extracts
	less than half the marginal coverage per extra sample---the signature of mass
	piled onto fewer modal solutions per task. This is \emph{predicted by}, not
	merely consistent with, the collapse literature: SFT on accuracy-selected targets
	concentrates probability on the modal solution even when the harvested targets
	are diverse \citep{dang2025weight,diffsmooth2025}. The rising
	\texttt{diversity\_mean} is therefore not in tension with the narrowing, because
	the two measure different things: \texttt{diversity\_mean} is the cardinality of
	the \emph{STaR-harvest} solution set during data generation, whereas the
	narrowing is in the eval-time pass@$K$ coverage of the trained model. Harvest
	diversity rises while the trained model's sampling coverage narrows---exactly the
	dissociation those accounts imply.
	
	\paragraph{Why headroom-paced, not collapse.} The per-round ceiling gain scales
	with how much headroom remains: it shrinks on the near-ceiling easy bank
	($+6.4 \to +2.3$) but holds on the low-base hard bank ($+3.2 \to +3.9$), and
	diversity rises throughout. Where the gains shrink, it is exhaustion of the
	\emph{reachable} frontier under a fixed base, not degeneration of the sampler. A
	frontier the base cannot reach is, by construction, invisible to a loop that can
	only train on what the verifier confirms---which is also why our setting, unlike
	prolonged RL with an exploration bonus \citep{liu2025prorl}, has no mechanism to
	manufacture compounding even in principle.
	
	\paragraph{Selector value erodes as the generator sharpens.} Divergence---the
	critic's only workspace---fell across the trajectory, the same dynamic reached
	from a different direction at the visible-rich end of the Stage-0 sweep. There is
	a practical lesson here for anyone designing such constellations: improving one
	role can shrink another role's niche, so the value of a learned selector is not
	static across a self-training run.
	
	\paragraph{What is and is not novel here.} We want to be candid about what is
	surprising. The crossover \emph{direction} is the expected one: under
	\citet{yue2025rlvr}, teacher-free training (no distillation) is predicted to stay
	base-bounded, because distillation is their only capability-importing
	intervention---so we do not claim the direction as a discovery. The contribution
	is threefold: (i)~confirming the effect in a regime that line has not
	examined---teacher-free iterated SFT with a free, total, exact verifier, where
	``reachable'' is measured exactly via large-$K$ pass@$K$; (ii)~the dissociation
	between rising harvest diversity and narrowing model coverage; and (iii)~the
	no-zero-frontier methodology, which shows that existence-of-compounding claims
	resting on zero-pass@$k$ cells must first establish those cells non-trivial at
	population scale.
	
	\paragraph{Scope.} One causal caveat deserves emphasis. The free, total, exact
	verifier is doing heavy lifting: it reinforces only correct-and-reachable
	programs and never explores, so ``no compounding'' may be a property of this
	\emph{verification regime} rather than of self-training in general. A noisy
	verifier, a learned reward model, or a teacher could behave differently. The
	result is, in that sense, domain-internal (\S\ref{sec:limits}).
	
	\section{Limitations (honesty bounds)}\label{sec:limits}
	\begin{itemize}[leftmargin=1.4em,itemsep=3pt]
		\item \textbf{Statistical power.} The gen\_v0 baseline, the Stage-1 loop, and
		the full $v_0/v_1/v_2$ ceiling trajectory are six-seed CI'd on both banks
		(eval-split variance of \emph{fixed} adapters). The no-acceleration claim
		additionally survives \emph{train-seed} variance: $K{=}4$ independent
		training trajectories decelerate on both banks (\S\ref{sec:stage2},
		Tier 3). The power limits that remain are honest ones: $K{=}4$ is small
		(the across-trajectory stdev is indicative, not a robust CI---$\geq 5$
		would be wanted), the train-seed replication covers the no-acceleration
		claim but not every per-round magnitude, and the weak-axis gradient
		(App.~\ref{app:weakaxis}) is still single-seed-per-bank.
		A $\geq 5$-seed train-seed CI for a robust across-trajectory interval is left to future work.
		\item \textbf{The decisive pass@$K$ measurement: the full crossover, at
			$K{=}64$, train-seed robust.} We measure both halves of the
		\citet{yue2025rlvr} signature at a matched $K{=}64$: $v_2$ wins at the
		operating budget (pass@8 $53.0\%/46.0\%$ vs base $46.0\%/38.0\%$) and the
		base overtakes $v_2$ at the large budget (pass@64 base $68.0\%/57.5\%$ vs
		$v_2$ $61.5\%/52.5\%$), on both banks. The overtaking holds across
		\emph{all four} independent training trajectories: every $v_2$ pass@64
		ceiling stays below the base's on both banks ($4/4$; easy margins
		$6$--$13$ tasks, hard $7$--$15$; \texttt{results/crossover\_tier3.json}),
		so the level test is train-seed robust and not a single-trajectory
		artifact. Two residual caveats: pass@$64$ lower-bounds $K\!\to\!\infty$
		(the true overtaking margin is at least this large), and the pass@$64$
		ceilings are single eval-seed---though the $+15.0$/$+11.5$~pp
		$v_2$@8-to-base@64 gaps dwarf the ${\sim}2$~pp eval-split noise, so
		eval-seed is not a fragile dependence. Per-point error bars (6-seed base
		and $v_2$ pass@$64$ CIs) are left to future work.
		\item \textbf{$n=1$ on most axes of generality:} one base model, one GPU, one
		DSL, one program prior. The results are domain-internal and support no
		external ``FlashFill capability'' claim.
		\item \textbf{Shared-base adapter attribution.} The generator and critic are
		LoRA adapters over the same frozen base, so they are not independent; a
		separately-based critic would be a cleaner test, and we have not done it.
		\item \textbf{Only two STaR rounds.} The no-acceleration claim (that per-round
		gain does not increase) rests on two paired steps, and the still-climbing
		hard bank could behave differently with more rounds. The crossover
		\emph{level} claim (that the trained model stays below the base at large
		$K$) is, by contrast, round-count-robust: it would survive additional
		rounds unless some later $v_N$ broke above the base envelope---which would
		itself be the compounding signal we are looking for.
		\item \textbf{No cross-domain benchmark.} All comparisons are internal-relative
		(critic vs.\ best-of-$k$, trained vs.\ base pass@$K$, both on our DSL).
		The baselines are instances of published methods
		(best-of-$n$/self-consistency, STaR, base pass@$K$) and are the correct
		referents for the questions asked (\S\ref{sec:related}), but we do not
		report accuracy on a standard PBE benchmark such as SyGuS, so the result
		is domain-internal.
	\end{itemize}
	
	\section{Reproducibility}\label{sec:repro}
	The evaluation banks are frozen and sha-pinned with tamper-evident metadata
	sidecars (easy \texttt{512ddf40\dots}, hard \texttt{49807fcc\dots}), and
	regeneration refuses to overwrite them without an explicit flag. Every RNG seed
	is recorded in each artifact's metadata, and the STaR data carries a mandatory
	frozen-bank leakage-exclusion guard that surfaces its reject counts. All analysis
	instruments are pure-stdlib and unit-tested---every instrument and the stdlib
	core carries a unit smoke test, all green---and the multi-seed aggregator prints
	every per-seed raw count and refuses to average over inconsistent task
	denominators. The stdlib core has zero third-party dependencies; the GPU stack is
	pinned behind a committed-lockfile \texttt{uv} extra.
	The code and all result artifacts are released at
	\url{https://github.com/izzortsi/trapdoor-loom} under the CC0 1.0
	public-domain dedication. The public repository de-anonymizes the authors, so
	this preprint is non-blind by design.
	
	\appendix
	\section{Full per-class tables}\label{app:perclass}
	
	Held-out solve@1/ceiling per structural class for gen\_v0/v1/v2 on both banks (single eval-split seed 1234). TOTALS equal the seed-1234 entries of the six-seed CIs (\S\ref{sec:stage2}) by construction: easy $51/92 \to 72/103 \to 76/106$, hard $45/76 \to 65/84 \to 66/92$ (solve@1/ceiling). These single-seed tables are diagnostic; the multi-seed ceiling trajectory of \S\ref{sec:stage2} carries the claim.
	
	\begin{table}[H]\centering\footnotesize
		\caption{Per-class held-out \textbf{solve@1/ceiling} on the \emph{easy} bank across STaR rounds (single eval-split seed 1234, $k{=}8$). Rows ordered by gen\_v0 support $n$; small-$n$ cells are suggestive only (\S\ref{sec:nozero}).}\label{tab:perclass-easy}
		\begin{tabular}{lcccc}
			\toprule
			class & $n$ & v0 & v1 & v2 \\
			\midrule
			a1{+}const & 34 & 17/28 & 26/31 & 25/29 \\
			a2{+}sub{+}const{+}match & 17 & 6/12 & 5/11 & 6/10 \\
			a1{+}sub{+}match & 14 & 0/5 & 5/8 & 5/9 \\
			a3{+}sub{+}const{+}match & 14 & 5/8 & 5/7 & 4/7 \\
			a3{+}sub{+}const{+}match{+}neg & 11 & 1/2 & 3/3 & 1/3 \\
			a3{+}sub{+}const{+}match{+}case & 9 & 5/6 & 4/6 & 5/6 \\
			a2{+}sub{+}const & 8 & 3/4 & 5/6 & 5/7 \\
			a3{+}sub{+}match{+}case{+}neg & 8 & 0/1 & 1/2 & 2/2 \\
			a2{+}sub{+}const{+}match{+}case & 7 & 3/4 & 3/5 & 3/5 \\
			a2{+}sub{+}const{+}match{+}neg & 7 & 1/2 & 1/2 & 1/2 \\
			a2{+}sub{+}match{+}neg & 7 & 0/1 & 1/2 & 1/2 \\
			a2{+}sub{+}match & 6 & 0/1 & 0/1 & 1/2 \\
			a2{+}sub{+}match{+}case{+}neg & 6 & 0/1 & 1/1 & 1/1 \\
			a3{+}sub{+}const{+}match{+}case{+}neg & 6 & 1/1 & 1/1 & 1/1 \\
			a1{+}sub{+}neg & 5 & 0/0 & 0/1 & 0/2 \\
			a1{+}sub{+}match{+}neg & 4 & 0/2 & 0/0 & 0/1 \\
			a2{+}const{+}case & 4 & 3/4 & 2/4 & 4/4 \\
			a1{+}sub{+}match{+}case & 3 & 0/0 & 0/0 & 0/0 \\
			a2{+}const & 3 & 2/3 & 3/3 & 3/3 \\
			a2{+}sub{+}const{+}match{+}case{+}neg & 3 & 0/0 & 0/0 & 0/0 \\
			a3{+}sub{+}match{+}neg & 3 & 0/0 & 0/1 & 0/0 \\
			a1{+}const{+}case & 2 & 0/2 & 0/0 & 0/2 \\
			a2{+}sub{+}const{+}neg & 2 & 0/0 & 0/0 & 0/0 \\
			a2{+}sub{+}match{+}case & 2 & 0/0 & 0/0 & 0/0 \\
			a3{+}sub{+}const & 2 & 1/2 & 2/2 & 2/2 \\
			a1{+}sub & 1 & 0/0 & 0/1 & 1/1 \\
			a1{+}sub{+}case & 1 & 0/0 & 0/0 & 1/1 \\
			a1{+}sub{+}match{+}case{+}neg & 1 & 0/0 & 0/0 & 0/0 \\
			a2{+}sub{+}const{+}case & 1 & 1/1 & 1/1 & 1/1 \\
			a2{+}sub{+}const{+}case{+}neg & 1 & 0/0 & 0/0 & 0/0 \\
			a2{+}sub{+}neg & 1 & 0/0 & 0/0 & 0/0 \\
			a3{+}const & 1 & 0/0 & 1/1 & 1/1 \\
			a3{+}const{+}case & 1 & 1/1 & 1/1 & 1/1 \\
			a3{+}sub{+}const{+}case & 1 & 1/1 & 1/1 & 1/1 \\
			a3{+}sub{+}const{+}neg & 1 & 0/0 & 0/0 & 0/0 \\
			a3{+}sub{+}match & 1 & 0/0 & 0/1 & 0/0 \\
			a3{+}sub{+}match{+}case & 1 & 0/0 & 0/0 & 0/0 \\
			a3{+}sub{+}neg & 1 & 0/0 & 0/0 & 0/0 \\
			\midrule
			TOTAL & 200 & 51/92 & 72/103 & 76/106 \\
			\bottomrule
		\end{tabular}
	\end{table}
	\noindent\textit{Easy bank, ceiling off-zero audit (\S\ref{sec:nozero}):} classes at $0$ ceiling in v0 that reach $>0$ by v2: \texttt{a1{+}sub{+}neg} ($n{=}5$, v2 ceil 2), \texttt{a1{+}sub} ($n{=}1$, v2 ceil 1), \texttt{a1{+}sub{+}case} ($n{=}1$, v2 ceil 1), \texttt{a3{+}const} ($n{=}1$, v2 ceil 1). Classes regressing to $0$: none. Every such cell has $n\le 5$, consistent with undersampling rather than an emergence/forgetting frontier.\par\medskip
	
	\begin{table}[H]\centering\footnotesize
		\caption{Per-class held-out \textbf{solve@1/ceiling} on the \emph{hard} bank across STaR rounds (single eval-split seed 1234, $k{=}8$). Rows ordered by gen\_v0 support $n$; small-$n$ cells are suggestive only (\S\ref{sec:nozero}).}\label{tab:perclass-hard}
		\begin{tabular}{lcccc}
			\toprule
			class & $n$ & v0 & v1 & v2 \\
			\midrule
			a1{+}const & 20 & 14/19 & 19/20 & 18/19 \\
			a3{+}sub{+}const{+}match & 16 & 2/6 & 5/7 & 4/6 \\
			a1{+}sub{+}match & 12 & 0/5 & 4/7 & 6/8 \\
			a4{+}sub{+}const{+}match{+}case{+}neg & 12 & 1/1 & 2/3 & 2/3 \\
			a2{+}sub{+}match{+}case{+}neg & 9 & 0/2 & 1/1 & 0/2 \\
			a3{+}sub{+}const{+}match{+}neg & 9 & 2/2 & 3/3 & 3/6 \\
			a4{+}sub{+}const{+}match{+}neg & 9 & 2/4 & 3/2 & 2/3 \\
			a2{+}sub{+}match & 8 & 2/2 & 2/2 & 2/3 \\
			a2{+}sub{+}match{+}neg & 8 & 1/1 & 1/2 & 1/1 \\
			a4{+}sub{+}const{+}match{+}case & 8 & 2/4 & 3/4 & 3/4 \\
			a2{+}sub{+}const{+}match & 7 & 2/4 & 2/3 & 2/2 \\
			a1{+}sub{+}match{+}neg & 6 & 0/2 & 2/4 & 3/4 \\
			a2{+}sub{+}const{+}match{+}case & 6 & 2/4 & 1/3 & 2/3 \\
			a3{+}sub{+}match{+}case{+}neg & 5 & 0/0 & 0/0 & 0/0 \\
			a1{+}sub{+}neg & 4 & 0/0 & 0/0 & 1/1 \\
			a2{+}const & 4 & 3/4 & 3/4 & 3/4 \\
			a2{+}sub{+}const & 4 & 2/1 & 1/1 & 1/2 \\
			a2{+}sub{+}const{+}neg & 4 & 1/2 & 1/2 & 1/3 \\
			a2{+}sub{+}match{+}case & 4 & 1/1 & 1/1 & 1/1 \\
			a4{+}sub{+}const{+}match & 4 & 1/1 & 1/2 & 1/1 \\
			a1{+}const{+}case & 3 & 1/1 & 2/2 & 2/3 \\
			a1{+}sub & 3 & 0/1 & 0/1 & 2/2 \\
			a2{+}const{+}case & 3 & 2/3 & 3/3 & 3/3 \\
			a3{+}sub{+}const & 3 & 0/0 & 0/1 & 0/1 \\
			a3{+}sub{+}const{+}match{+}case & 3 & 0/1 & 0/0 & 0/0 \\
			a3{+}sub{+}match{+}neg & 3 & 0/0 & 0/0 & 0/0 \\
			a1{+}sub{+}match{+}case & 2 & 0/0 & 0/0 & 0/0 \\
			a2{+}sub{+}const{+}case & 2 & 0/0 & 1/1 & 1/2 \\
			a3{+}const & 2 & 2/2 & 2/2 & 1/2 \\
			a3{+}sub{+}const{+}match{+}case{+}neg & 2 & 0/1 & 0/1 & 0/1 \\
			a3{+}sub{+}const{+}neg & 2 & 0/0 & 0/0 & 0/0 \\
			a3{+}sub{+}match & 2 & 0/0 & 0/0 & 0/0 \\
			a4{+}sub{+}match{+}case & 2 & 0/0 & 0/0 & 0/0 \\
			a2{+}sub{+}case{+}neg & 1 & 0/0 & 0/0 & 0/0 \\
			a2{+}sub{+}const{+}case{+}neg & 1 & 0/0 & 0/0 & 0/0 \\
			a2{+}sub{+}const{+}match{+}case{+}neg & 1 & 0/0 & 0/0 & 0/0 \\
			a2{+}sub{+}const{+}match{+}neg & 1 & 0/0 & 0/0 & 0/0 \\
			a3{+}sub{+}const{+}case & 1 & 0/0 & 0/0 & 0/0 \\
			a3{+}sub{+}const{+}case{+}neg & 1 & 0/0 & 0/0 & 0/0 \\
			a4{+}sub{+}const & 1 & 1/1 & 1/1 & 0/1 \\
			a4{+}sub{+}const{+}case & 1 & 1/1 & 1/1 & 1/1 \\
			a4{+}sub{+}match{+}neg & 1 & 0/0 & 0/0 & 0/0 \\
			\midrule
			TOTAL & 200 & 45/76 & 65/84 & 66/92 \\
			\bottomrule
		\end{tabular}
	\end{table}
	\noindent\textit{Hard bank, ceiling off-zero audit (\S\ref{sec:nozero}):} classes at $0$ ceiling in v0 that reach $>0$ by v2: \texttt{a1{+}sub{+}neg} ($n{=}4$, v2 ceil 1), \texttt{a3{+}sub{+}const} ($n{=}3$, v2 ceil 1), \texttt{a2{+}sub{+}const{+}case} ($n{=}2$, v2 ceil 2). Classes regressing to $0$: \texttt{a3{+}sub{+}const{+}match{+}case} ($n{=}3$, v0 ceil 1). Every such cell has $n\le 5$, consistent with undersampling rather than an emergence/forgetting frontier.\par\medskip
	
	\section{Prompt formats and chat templating}\label{app:prompts}
	All roles share one blessed formatting path (\texttt{chatfmt}, \texttt{criticfmt},
	\texttt{traindata.format\_prompt}); training and inference call the same
	functions, so the surface cannot drift. Inputs and outputs are rendered with
	Python \texttt{repr()} so whitespace is exact and never silently stripped. The
	Qwen3 wrapping below is the committed \texttt{chat\_template.jinja} (saved beside
	each adapter); the operator's \texttt{probe\_api.sh} renders byte-identical
	strings from the live tokenizer. The template emits \emph{no} default system
	turn when no system message is supplied, and \texttt{eos\_token} is
	\texttt{<|im\_end|>}.
	
	\paragraph{Generator --- generation prompt} (\texttt{visible examples} $\to$
	\texttt{program}; ends at \texttt{assistant\textbackslash n}, where sampling
	begins):
	\begin{lstlisting}
		<|im_start|>user
		Infer the string transform program from these examples:
		'Zoe Lee' => 'L'
		'Al Roy' => 'R'
		Program:<|im_end|>
		<|im_start|>assistant
	\end{lstlisting}
	
	\paragraph{Generator --- full training row} (user turn $+$ assistant
	completion). With \texttt{completion\_only\_loss=True} the loss covers only the
	completion span---the program text plus \texttt{<|im\_end|>}---and the prompt is
	masked, exactly as the \texttt{train\_generator.py --dry-run} gate prints
	(\texttt{SUPERVISED = program$+$eos}, \texttt{MASKED = prompt}):
	\begin{lstlisting}
		<|im_start|>user
		Infer the string transform program from these examples:
		'Zoe Lee' => 'L'
		'Al Roy' => 'R'
		Program:<|im_end|>
		<|im_start|>assistant
		Concat([Case(upper,SubStr(MatchPos(WORD,-1,start),CPos(-1)))])<|im_end|>
	\end{lstlisting}
	
	\paragraph{Critic --- generation prompt} ($(\text{examples},\text{candidate})
	\to \text{yes/no}$; the candidate already fits the visible examples, so the
	question is held-out generalization). At inference the survivors are ranked by
	the $P(\texttt{yes})-P(\texttt{no})$ first-token logit margin:
	\begin{lstlisting}
		<|im_start|>user
		Given these input/output examples:
		'Zoe Lee' => 'L'
		'Al Roy' => 'R'
		and this candidate program: Concat([SubStr(CPos(-1),CPos(-1))])
		Does the program correctly capture the transform for unseen inputs of the same kind? Answer 'yes' or 'no'.<|im_end|>
		<|im_start|>assistant
	\end{lstlisting}
	The critic completion is \texttt{yes<|im\_end|>} or \texttt{no<|im\_end|>}. The
	critic prompt need only be internally consistent (train $=$ inference); it does
	not, and need not, match the generator's surface.
	\section{Single-seed weak-axis gradient (demoted from main text)}\label{app:weakaxis}
	This single-seed (seed 1234) structural decomposition lives in the appendix
	because, unlike the six-seed trajectory of \S\ref{sec:stage2}, it is not
	multi-seeded, and its round-1 pattern does not replicate in round 2.
	
	\begin{table}[h]\centering
		\caption{Headroom-normalized fraction closed per axis (single seed 1234).}
		\begin{tabular}{lcccc}
			\toprule
			& R1 easy & R1 hard & R2 easy & R2 hard \\
			\midrule
			\textbf{match-no-const (weak)} & \textbf{11.1\%} & \textbf{8.2\%} & 2.5\% (1-task) & 4.4\% \\
			const-anchored (easy)          & 7.4\% & 5.9\% & 0.0\% & 6.2\% \\
			neg-bearing (control)          & 5.3\% & 4.7\% & 1.9\% & 9.8\% \\
			weak\,:\,easy ratio            & 1.50$\times$ & 1.39$\times$ & --- & inverts \\
			\bottomrule
		\end{tabular}
	\end{table}
	
	In round~1 the weak axis closed the largest headroom fraction on both banks,
	above both the easy axis and the low-base \texttt{neg} control---``improved most
	where weakest.'' In round~2 the preference vanishes: a 1-task noise move on easy,
	and an inversion on hard. The preferentially-improved axis is thus a one-round
	transient. Because this is a single-seed-per-bank metric rather than a
	multi-seeded claim, it does not conflict with the six-seed ceiling trajectory of
	\S\ref{sec:stage2}.
	
	\bibliographystyle{plainnat}
	\bibliography{refs}
	
\end{document}